%% file: main.tex
\documentclass[runningheads]{llncs}

\usepackage{eccv}

\usepackage{eccvabbrv}

\usepackage[accsupp]{axessibility}  %

\usepackage{orcidlink}
\usepackage{graphicx}
\usepackage{booktabs}
\usepackage{amsmath}
\usepackage{amssymb}
\usepackage{multirow}
\usepackage{xcolor}
\usepackage{tabularx}
\usepackage{colortbl}  %
\usepackage{array}
\usepackage{siunitx} %
\usepackage{tcolorbox}  %
\usepackage{listings}
\usepackage{mathrsfs}
\usepackage{wrapfig}
\usepackage{geometry}
\usepackage{upgreek}
\usepackage{makecell}

\usepackage{mwe} %
\usepackage{calc} %
\usepackage{pifont} %

\usepackage{marvosym}   %

\usepackage{hyperref}

\newcommand{\yes}{\color{blue}{\ding{51}}}
\newcommand{\no}{\color{red}{\ding{55}}}
\definecolor{Gray}{gray}{0.85}
\newcommand{\method}{\mbox{{ManiGaussian}}\xspace}

\definecolor{best}{rgb}{0.96, 0.57, 0.58}
\definecolor{second}{rgb}{0.98, 0.78, 0.57}
\definecolor{third}{rgb}{1.0, 1.0, 0.56}

\newcommand{\dd}[1]{$#1$}
\newcommand{\ddbf}[1]{$\mathbf{#1}$}
\newcommand{\ddbfblue}[1]{\underline{$#1$}}

\let\titleold\title
\renewcommand{\title}[1]{\titleold{#1}\newcommand{\thetitle}{#1}}
\def\maketitlesupplementary
   {
    {\newpage
        \centering
        \Large
        \textbf{\thetitle}\\
        Supplementary Material \\
        }
   }

\begin{document}

\title{ManiGaussian: Dynamic Gaussian Splatting for Multi-task Robotic Manipulation}

\author{Guanxing Lu\inst{1}\orcidlink{0000-0002-6869-7647} \and Shiyi Zhang\inst{1}\orcidlink{0009-0005-8067-4025} \and
Ziwei Wang\inst{2,3}\textsuperscript{\Letter}\thanks{\Letter~ Corresponding author.}\orcidlink{0000-0001-9225-8495} \and
Changliu Liu\inst{4}\orcidlink{0000-0002-3767-5517},\\
Jiwen Lu\inst{2}\orcidlink{0000-0002-6121-5529} \and
Yansong Tang\inst{1}\orcidlink{0000-0002-1534-4549
}}

\authorrunning{G.~Lu et al.}

\titlerunning{ManiGaussian}

\institute{Shenzhen Key Laboratory of Ubiquitous Data Enabling, Shenzhen International Graduate School, Tsinghua University
\and
Department of Automation, Tsinghua University
\and
Nanyang Technological University, $^4$ Carnegie Mellon University
\\
\email{\{lgx23@mails.,sy-zhang23@mails.,lujiwen@,tang.yansong@sz.\}tsinghua.edu.cn}\\
\email{\{ziweiwa2@,cliu6@\}andrew.cmu.edu}
}

\maketitle
\input{sec/0_abstract}

\input{sec/1_introduction}

\input{sec/2_related_work}

\input{sec/3_approach}

\input{sec/4_experiments}
\input{sec/5_conclusion}

\bibliographystyle{splncs04}
% \bibliography{egbib}

\input{sec/X_suppl}

\end{document}

%% file: sec/0_abstract.tex
\begin{abstract}

Performing language-conditioned robotic manipulation tasks in unstructured environments is highly demanded for general intelligent robots. Conventional robotic manipulation methods usually learn a semantic representation of the observation for action prediction, which ignores the scene-level spatiotemporal dynamics for human goal completion. In this paper, we propose a dynamic Gaussian Splatting method named \textbf{ManiGaussian} for multi-task robotic manipulation, which mines scene dynamics via future scene reconstruction. Specifically, we first formulate the dynamic Gaussian Splatting framework that infers the semantics propagation in the Gaussian embedding space, where the semantic representation is leveraged to predict the optimal robot action. Then, we build a Gaussian world model to parameterize the distribution in our dynamic Gaussian Splatting framework, which provides informative supervision in the interactive environment via future scene reconstruction. We evaluate our ManiGaussian on \num{10} RLBench tasks with \num{166} variations, and the results demonstrate our framework can outperform the state-of-the-art methods by \num{13.1}\% in average success rate\footnote[1]{Project page: \url{https://guanxinglu.github.io/ManiGaussian/}}.


\end{abstract}

%% file: sec/1_introduction.tex
\section{Introduction}
\label{sec:intro}

Designing autonomous agents for language-conditioned manipulation tasks~\cite{tenorth2010understanding,tellex2011understanding,kalashnikov2018qtopt,brohan2022rt1,jang2022bcz,zeng2021transporter,shridhar2022cliport,shridhar2023peract,fu2024mobile} has been highly desired in the pursuit of artificial intelligence for a long time. In realistic deployment, intelligent robots are usually required to deal with unseen scenarios in novel tasks. Therefore, comprehending complex 3D structures in the deployment scenes is necessary for the robots to achieve high task success rates across diverse manipulation tasks.

\begin{figure}[t]
    \centering
    \includegraphics[width=1\textwidth]{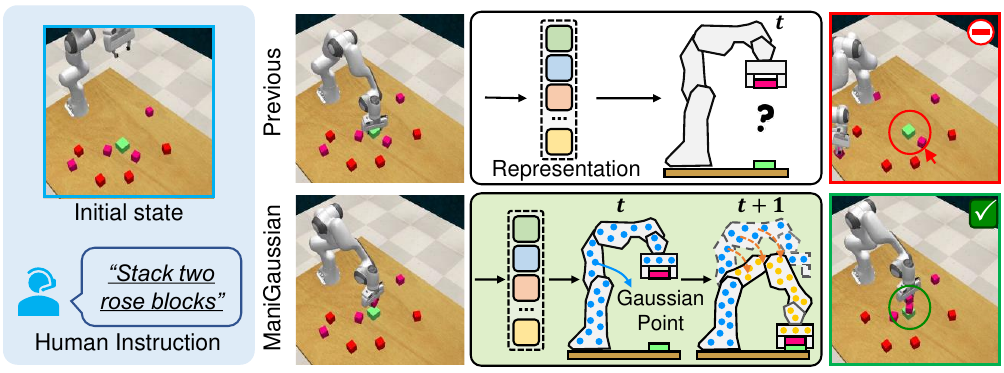}
    \caption{\small 
    Consider the human instruction \emph{``stack two rose blocks''}, where the task is considered successful if two rose blocks are stacked upon the green block.
    The previous method (GNFactor~\cite{ze2023gnfactor}) attempts to pick up the fixed green base but fails severely due to the misunderstanding of the scene dynamics, while our \method completes the task successfully by explicitly encoding the scene dynamics via future scene reconstruction in Gaussian embedding space.
    }
    \label{fig:teaser}
\end{figure}

To address the challenges, previous arts have made great progress in general manipulation policy learning, which can be divided into two categories including perceptive methods and generative methods. 
For the first regard, semantic features extracted by perceptive models are directly leveraged to predict the robot actions according to the visual input such as image~\cite{liu2022instruction,guhur2023instruction,goyal2023rvt}, point cloud~\cite{chen2023polarnet,gervet2023act3d,zhang2023universal} and voxel~\cite{james2022coarse,shridhar2023peract}. 
However, the perceptive methods heavily rely on multi-view or gripper-mounted cameras to cover the whole workbench to deal with the occlusion problem within unstructured environments, which restricts their deployment.
To this end, generative methods~\cite{parisi2022pvr,nair2022r3m,radosavovic2023realmvp,laskin2020curl,hansen2021soda,ze2023rl3d,ze2023gnfactor,li20223d} capture the 3D scene structure information by reconstructing the scene and objects in arbitrary novel views with self-supervised learning. Nevertheless, they ignore the spatiotemporal dynamics that depict the physical interaction among objects during manipulation, and the predicted actions still fail to complete human goals without correct object interactions. \Cref{fig:teaser} shows a comparison of manipulation achieved by the conventional generative manipulation method (top) and the proposed method (bottom), where the conventional method fails to stack the two rose blocks due to the poor comprehension of scene dynamics.

In this paper, we propose a \method method that leverages a dynamic Gassuain Splatting framework for multi-task robotic manipulation. Different from conventional methods which only focus on semantic representation, our method mines the scene-level spatiotemporal dynamics via future scene reconstruction. Therefore, the interaction among objects can be comprehended for accurate manipulation action prediction. More specifically, we first formulate the dynamic Gaussian Splatting framework that models the propagation of diverse semantic features in the Gaussian embedding space, and the semantic features with scene dynamics are leveraged to predict the optimal robot actions for general manipulation tasks. We build a Gaussian world model to parameterize the distributions in our dynamic Gaussian Splatting framework. Therefore, our framework can acquire informative supervision in interactive environments by reconstructing the future scene according to the current scene and the robot actions, where we constrain consistency between reconstructed and realistic future scenes for dynamics mining. We evaluate our \method method on the RLBench dataset~\cite{james2020rlbench} with \num{10} tasks and \num{166} variants, where our method outperforms the state-of-the-art multi-task robotic manipulation methods by \num{13.1}\% in the average task success rate. Our contributions can be summarized as follows:

\begin{itemize}
    \item We propose a dynamic Gaussian Splatting framework to learn the scene-level spatiotemporal dynamics in general robotic manipulation tasks, so that the robotic agent can complete human instructions with accurate action prediction in unstructured environments.

    \item We build a Gaussian world model to parameterize distributions in our dynamic Gaussian Splatting framework, which can provide informative supervision to learn scene dynamics from the interactive environment.

    \item We conduct extensive experiments of \num{10} tasks on RLBench, and the results demonstrate that our method achieves a higher success rate than the state-of-the-art methods with less computation.
    
\end{itemize}

%% file: sec/2_related_work.tex
\section{Related Work}
\label{sec:related_work}

\noindent\textbf{Visual Representations for Robotic Manipulation.} 
Developing intelligent agents for language-conditioned manipulation tasks in complex and unstructured environments has been a longstanding objective.
One of the key bottlenecks in achieving this goal is effectively representing visual information of the scene.
Prior arts can be categorized into two branches: perceptive methods and generative methods. Perceptive methods directly utilize pretrained 2D~\cite{liu2022instruction,guhur2023instruction,goyal2023rvt,zhang2023universal} or 3D visual representation backbone~\cite{chen2023polarnet,gervet2023act3d,james2022coarse,shridhar2023peract} to learn scene embedding, where optimal robot actions are predicted based on the scene semantics.
For example, InstructRL~\cite{liu2022instruction} and Hiveformer~\cite{guhur2023instruction} directly passed 2D visual tokens through a multi-modal transformer to decode gripper actions, but struggled to handle complex manipulation tasks due to the lack of geometric understanding.
To incorporate 3D information beyond images, PolarNet~\cite{chen2023polarnet} and Act3D~\cite{gervet2023act3d} utilized point cloud representation, where PolarNet used a PointNeXt~\cite{qian2022pointnext}-based architecture and Act3D designed a ghost point sampling mechanism to decode actions.
Moreover, PerAct~\cite{shridhar2023peract} fed voxel tokens into a PerceiverIO~\cite{jaegle2021perceiver}-based transformer policy, demonstrating impressive performance in a variety of manipulation tasks.
However, perceptive methods heavily rely on seamless camera overlay for comprehensive 3D understanding, which makes them less effective in unstructured environments.
To address this, generative methods~\cite{parisi2022pvr,nair2022r3m,radosavovic2023realmvp,laskin2020curl,hansen2021soda,ze2023rl3d,ze2023gnfactor,li20223d} have gained attention. which learns the 3D geometry through self-supervised novel view reconstruction.
For instance, Li \etal~\cite{li20223d} combined NeRF and time contrastive learning to embed 3D geometry and learn fluid dynamics within an autoencoder framework.
GNFactor~\cite{ze2023gnfactor} optimized a generalizable NeRF with a reconstruction loss besides behavior cloning, and showed effective improvement in both simulated and real scenarios.
However, conventional generative methods usually ignore the scene-level spatiotemporal dynamics that demonstrate the interaction among objects, and the predicted actions still fail to achieve human goals because of incorrect interaction.

\noindent\textbf{World Models.} 
In recent years, world models have emerged as an effective approach to encode scene dynamics by predicting the future states given the current state and actions, which are explored in autonomous driving~\cite{wang2023drivedreamer,gao2022enhance,hu2022model,hu2023gaia}, game agent~\cite{ha2018recurrent,hafner2022deep,hafner2019dream,hafner2020mastering,hafner2023mastering,ye2021mastering,schrittwieser2020mastering} and robotic manipulation \cite{hansen2023td,wu2023daydreamer,seo2023masked}. Early works~\cite{ha2018recurrent,hafner2022deep,hafner2019dream,hafner2020mastering,hafner2023mastering,hansen2023td,ye2021mastering,schrittwieser2020mastering,seo2023masked} learned a latent space for future prediction by autoencoding, which acquired notable effectiveness in both simulated and real-world situations~\cite{wu2023daydreamer}. However, learning latent for accurate future prediction requires a large amount of data and is limited to simple tasks such as robot control due to the weak representative ability of the implicit features. To address these limitations, explicit representation in the image domain~\cite{du2024learning,seo2022reinforcement,wu2024pre,mendonca2023structured} and the language domain~\cite{lin2023learning,lu2023thinkbot,chen2024s,wang2023describe,hong2024data} has been widely studied because of the rich semantics.
UniPi~\cite{du2024learning} reconstructed the future images with a text-conditional video generation model, employing an inverse dynamics model to obtain the intermediate actions. Dynalang~\cite{lin2023learning} learned to predict text representations as future states, and enabled embodied agents to navigate in photorealistic home scans under human instructions.
In contrast to these approaches, we generalize the world model to embedding space of dynamic Gaussian Splatting, which predicts the future state for the agent to learn scene-level dynamics from interactive environments.

\noindent\textbf{Gaussian Splatting.} 
Gaussian Splatting~\cite{kerbl20233d} models the scenes with a set of 3D Gaussians which are projected to 2D planes with efficient differentiable splatting. Gaussian Splatting achieves higher effectiveness and efficiency compared with implicit representations such as Neural Radiance Fields (NeRF)~\cite{mildenhall2021nerf,jiang2021giga,lin2023mira,li20223d,driess2022nerfrl,shim2023snerl,ze2023gnfactor} with fast inference, high fidelity, and strong editability for novel view synthesis. Please refer to~\cite{chen2024survey} for a comprehensive survey on 3D Gaussian Splatting. To deploy Gaussian Splatting in diverse complex scenarios, many variants have been proposed to enhance the generalization ability, enrich the semantic information, and reconstruct deformable scenes. 
For higher generalization ability across diverse scenes, recent works~\cite{szymanowicz2023splatter, zheng2023gps,zou2023triplane,charatan2023pixelsplat,fu2023colmap,xu2024agg,zhang2024gaussiancube} constructed a direct mapping from pixels to Gaussian parameters from large-scale datasets.
To integrate rich semantic information into Gaussian Splatting, many efforts  \cite{qin2023langsplat,zuo2024fmgs,shi2023language,zhou2024feature} have been demonstrated in distilling Gaussian radiance fields from pretrained foundation models~\cite{radford2021clip,caron2021dino,rombach2022diffusion,kirillov2023segment}.
For instance, LangSplat~\cite{qin2023langsplat} advanced the Gaussian representation by encoding language features distilled from CLIP~\cite{radford2021clip} using a scene-wise language autoencoder, enabling efficient open-vocabulary localization compared with its NeRF-based counterpart~\cite{kerr2023lerf}.
For deformation modeling, time-variant Gaussian radiance fields 
\cite{yang2023real,xie2023physgaussian,yang2023deformable,luiten2023dynamic,wu20234d,liang2023robo360,abou2023physically} were reconstructed from videos instead of images, which are widely applied in applications such as surgical scene reconstruction~\cite{zhu2024deformable,liu2024endogaussian}.
Although these approaches have achieved high-quality reconstruction from entire videos like interpolation, extrapolation to future states conditioned on previous states and actions is unexplored, which holds significance for scene-level dynamics modeling for interactive agents.
In this paper, we formulate a dynamic Gaussian Splatting framework to model the scene dynamics of object interactions, which enhances the physical reasoning for agents to complete a wide range of robotic manipulation tasks.

%% file: sec/3_approach.tex
\section{Approach}
\label{sec:approach}

In this section, we first briefly introduce preliminaries on the problem formulation (\Cref{subsec:prelimilaries}), and then we present an overview of our pipeline (\Cref{subsec:overall}).
Subsequently, we introduce a dynamic Gaussian Splatting framework (\Cref{subsec:dynamic_gaussian_splatting}) that infers the propagation semantics of the manipulation scenarios in the Gaussian embedding space. To enable our dynamic Gaussian Splatting framework to learn scene dynamics from the interactive environment, we build a Gaussian world model (\Cref{subsec:gaussian_world_model}) that reconstructs future scenes according to the propagated semantics.

\subsection{Problem Formulation}\label{subsec:prelimilaries}

The demand for language-conditioned robotic manipulation is a significant aspect in the development of general intelligent robots.
The agent is required to interactively predict the subsequent pose of the robot arm based on the observation and achieve the pose with a low-level motion planner to complete a wide range of manipulation tasks described in humans.
The visual input at the $t_{th}$ step for the agent is defined as $o^{(t)}=(\mathbf{C}^{(t)}, \mathbf{D}^{(t)}, \mathbf{P}^{(t)})$, where $\mathbf{C}^{(t)}$ and $\mathbf{D}^{(t)}$ respectively represent the single-view images and the depth images. The proprioception matrix $\mathbf{P}^{(t)}\in \mathbb{R}^4$ indicates the gripper state including the end-effector position, openness, and current timestep. Based on the visual input $o^{(t)}$ and the language instructions, the agent is required to generate the optimal action for the robot arm and grippers $\mathbf{a}^{(t)}=(\mathbf{a}^{(t)}_{\text{trans}},\ \mathbf{a}^{(t)}_{\text{{rot\phantom{}}}},\ \mathbf{a}^{(t)}_{\text{{open\phantom{t}}}}, \mathbf{a}^{(t)}_{\text{col}})$, which respectively demonstrates the target translation in voxel $\mathbf{a}^{(t)}_{\text{trans}}\!\in\! \mathbb{R}^{\!100^3}$, rotation $\mathbf{a}_{\text {rot}}^{(t)}\!\in\! \mathbb{R}^{\!(360\! /\! 5) \!\times\! 3}$, openness $\mathbf{a}^{(t)}_{\text{{open\phantom{t}}}}\!\in\! [0,\!1\!]$ and collision avoidance $\mathbf{a}^{(t)}_{\text{col}}\!\in\! [0,\!1\!]$. 

To learn the manipulation policy effectively, expert demonstrations as offline datasets are provided for imitation learning, where the sample triplets contain the visual input, language instruction and expert actions. Existing methods leverage powerful visual representations to learn informative latent features for optimal action prediction.
However, they ignore the spatiotemporal dynamics which depicts the physical interaction among objects, and the predicted actions usually fail to complete complex human goals without correct object interactions. On the contrary, we present a dynamic Gaussian Splatting framework to mine the scene dynamics for robotic manipulation.

\begin{figure}[t]
    \centering
    \includegraphics[width=1\textwidth]{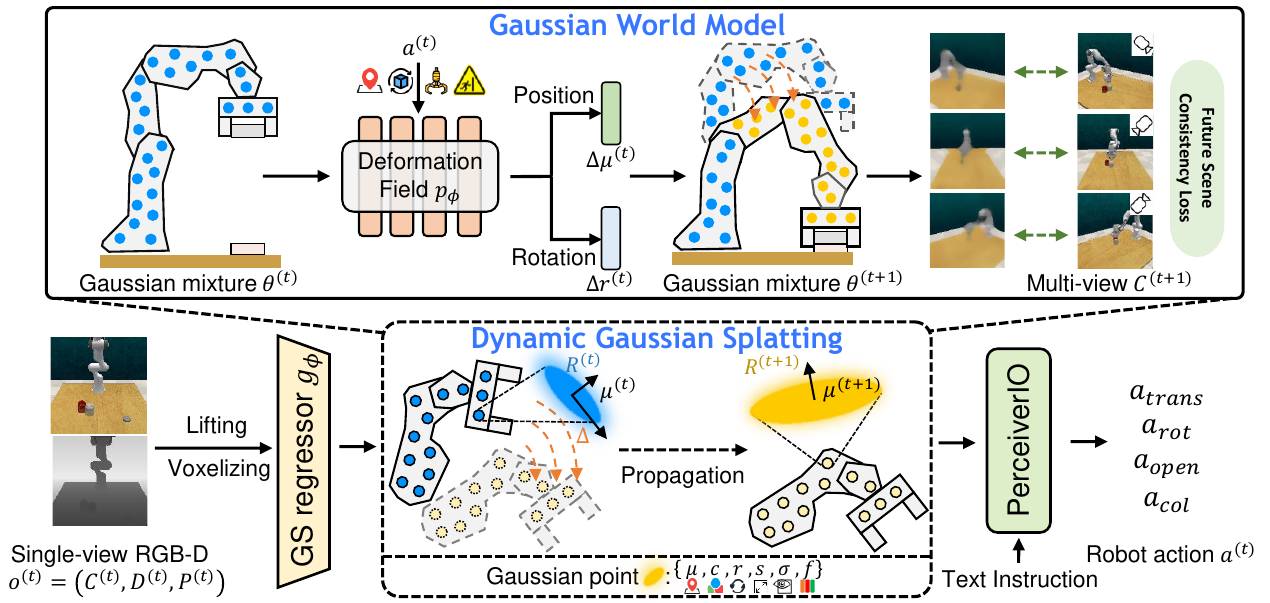}
    \caption{\small The overall pipeline of \method, which primarily consists of a dynamic Gaussian Splatting framework and a Gaussian world model. The dynamic Gaussian Splatting framework models the propagation of diverse semantic features in the Gaussian embedding space for manipulation, and the Gaussian world model parameterizes distributions to provide supervision by reconstructing the future scene for scene-level dynamics mining.
    }
    \label{fig:pipeline}
\end{figure}

\subsection{Overall Pipeline}\label{subsec:overall}

The overall pipeline of our \method method is shown in \Cref{fig:pipeline}, in which we construct a dynamic Gaussian Splatting framework that models the propagation of diverse semantic features in the Gaussian embedding space for manipulation. We also build a Gaussian world model to parameterize distributions in our dynamic Gaussian Splatting framework, which can provide informative supervision of scene dynamics by future scene reconstruction.
More specifically, we transform the visual input from RGB-D cameras to a volumetric representation by lifting and voxelization for data preprocessing. For dynamic Gaussian Splatting, we leverage a Gaussian regressor to infer the Gaussian distribution of geometric and semantic features in the scene based on the representation, propagated along time steps with rich scene-level spatiotemporal dynamics. For the Gaussian world model, we instantiate a deformation field to reconstruct the future scene according to the current scene and the robot actions, and require consistency between reconstructed and realistic scenes for dynamics mining. Therefore, the spatiotemporal dynamics indicating object correlation can be embedded into the representation learned in the dynamic Gaussian Splatting framework. Finally, we employ multi-modal transformer PerceiverIO~\cite{jaegle2021perceiver} to predict the optimal robot actions for general manipulation tasks, which considers geometric, semantic, and dynamic information with human instructions.

\subsection{Dynamic Gaussian Splatting for Robotic Manipulation}\label{subsec:dynamic_gaussian_splatting}
In order to capture the scene-level dynamics for general manipulation tasks, we propose a dynamic Gaussian Splatting framework that models the propagation of diverse semantic features within the Gaussian embedding space.
While the vanilla Gaussian Splatting has remarkable effectiveness and efficiency in reconstructing static environments, it fails to capture the scene dynamics for manipulation due to the lack of temporal information.
To this end, we formulate a dynamic Gaussian Splatting framework based on the vanilla Gaussian Splatting methodology by enabling the Gaussian points of scene representation to move with robotic manipulation, which demonstrates the physical interactions between objects. The scene representation contains geometric information depicting the explicit visual clues, semantic information illustrating the implicit high-level visual features, and dynamic information encoding the physical properties of the scene, which are utilized to predict the optimal actions.

\noindent\textbf{Dynamic Gaussian Splatting.}
Gaussian Splatting~\cite{kerbl20233d} is a promising approach for multi-view 3D reconstruction, which exhibits fast inference, high fidelity, and strong editability of generated content compared with Neural Radiance Field (NeRF)~\cite{mildenhall2021nerf}. Gaussian Splatting represents a 3D scene explicitly with multiple Gaussian primitives,
where the $i_{th}$ Gaussian primitive is parameterized by $\theta_i = (\mu_i, c_i, r_i, s_i, \sigma_i)$, where respectively represent the positions, color, rotation, scale, and opacity for the Gaussian primitive.
To render a novel view, we project Gaussian primitives onto the 2D plane by differential tile-based rasterization. The value of the pixel $\mathbf{p}$ can be rendered by the alpha-blend rendering:
\begin{equation}\label{eq:rgb_rendering}
    C(\mathbf{p}) =\! \sum_{i=1}^{N} \alpha_i c_i \prod_{j=1}^{i-1} (1-\alpha_j) \quad \text { where, } \alpha_i=\sigma_i e^{-\frac{1}{2}\left(\mathbf{p}-\mu_i\right)^{\top} \Sigma_i^{-1}\left(\mathbf{p}-\mu_i\right)},
\end{equation}
where $C$ is the rendered image, $N$ denotes the number of Gaussians in this tile, $\alpha_i$ represents the 2D density of the Gaussian points in the splatting process, and $\Sigma_i$ stands for the covariance matrix acquired from the rotation and scales of the Gaussian parameters.
However, the vanilla Gaussian Splatting encounters difficulties in reconstructing changing environments, which limits the ability to model scene-level dynamics that is crucial for manipulation tasks. 
To address this, we enable the Gaussian particles to be propagated with time to capture the spatiotemporal dynamics of the scene. The parameters of the $i_{th}$ Gaussian primitive at the $t_{th}$ step can be expressed as follows:
\begin{equation}\label{eq:dynamic_gaussian_Splatting}
    \theta_i^{(t)}=(\mu_{i}^{(t)}, c_i^{(t)}, r_{i}^{(t)}, s_i^{(t)}, \sigma_i^{(t)}, f_i^{(t)}).
\end{equation}
The positions, colors, rotations, scales, and opacities with the superscript $t$ represent their counterparts at the $t_{th}$ step in the propagation, and $f_i^{(t)}$ is the high-level semantic feature distilled from the Stable Diffusion~\cite{rombach2022diffusion} visual encoder based on the RGB images of the scene. In robotic manipulation, all objects are regarded as rigid bodies without inherent properties including colors, scales, opacities, and semantic features. $c_i^{(t)}$, $s_i^{(t)}$, $\sigma_i^{(t)}$ and $f_i^{(t)}$ are therefore regarded as time-independent parameters. The positions and rotation of Gaussian particles change during the manipulation due to the physical interaction between objects and robot grippers, which can be formulated as follows:
\begin{equation}\label{eq:propagation}
    (\mu_{i}^{(t+1)}, r_{i}^{(t+1)})=(\mu_{i}^{(t)}+\Updelta \mu_{i}^{(t)}, r_{i}^{(t)} + \Updelta r_{i}^{(t)})
\end{equation}where $\Updelta \mu_{i}^{(t)}$ and $\Updelta r_{i}^{(t)}$ demonstrate the change of positions and rotation from the $t_{th}$ step to the next one for the $i_{th}$ Gaussian primitive. With the time-dependent parameters of the Gaussian mixture distribution, the pixel values in 2D views of the scene can still be rendered by (\ref{eq:rgb_rendering}).

\noindent\textbf{Gaussian World Model.} In our implementation, we present a Gaussian world model to parameterize the Gaussian mixture distribution in dynamic Gaussian Splatting, through which the future scene can be reconstructed via parameter propagation. Therefore, the dynamic Gaussian Splatting model can acquire informative supervision in the interactive environment by considering the consistency between the reconstructed and realistic feature scenes. World models are effective to learn the environmental dynamics for downstream tasks by anticipating the future state $s^{(t+1)}$ based on the current state $s^{(t)}$ and action $a^{(t)}$ at the $t_{th}$ step, which have been applied to a variety of tasks including autonomous driving~\cite{wang2023drivedreamer,gao2022enhance,hu2022model,hu2023gaia} and game agent~\cite{ha2018recurrent,hafner2022deep,hafner2019dream,hafner2020mastering,hafner2023mastering,ye2021mastering,schrittwieser2020mastering}.
For our robotic manipulation tasks, we instantiate the current state in the world model as the visual observation in the current step, and actions refer to those of the robot arm and grippers. They are leveraged to predict the visual scenes observed in the next step that represent the future state. More specifically, the Gaussian world model contains a representation network $q_{\phi}$ that learns high-level visual features with rich semantics for the input observation, a Gaussian regressor $g_{\phi}$ that predicts the Gaussian parameters of different primitives based on the visual features, a deformation predictor $p_{\phi}$ that infers the difference of Gaussian parameters during the propagation, and a Gaussian renderer $\mathcal{R}$ (\ref{eq:rgb_rendering}) that generates the RGB images for the predicted future state:
\begin{equation}\label{eq:gwm}
\begin{cases}
\text{Representation model:} 
& \mathbf{v}^{(t)} = q_\phi\left(o^{(t)}\right), \\ 
\text{Gaussian regressor: } 
& \theta^{(t)} = g_\phi\left(\mathbf{v}^{(t)}\right), \\ 
\text{Deformation predictor:} 
& \Updelta \theta^{(t)} = p_\phi\left(\theta^{(t)}, a^{(t)}\right), \\ 
\text{Gaussian renderer: } 
& o^{(t+1)} = \mathcal{R}\left(\theta^{(t+1)}, w\right), \\  
\end{cases}
\end{equation}where $o^{(t)}$ and $\mathbf{v}^{(t)}$ mean the visual observation and the corresponding high-level visual features at the $t_{th}$ step. $w$ is the camera pose for the view where we project the Gaussian primitives. We leverage multi-head neural networks as the Gaussian regressor, where each head predicts a specific feature for the Gaussian parameters shown in (\ref{eq:dynamic_gaussian_Splatting}). By inferring the changes of positions and rotations between consecutive steps, we acquire the propagated Gaussian parameters in the future step based on (\ref{eq:propagation}). Finally, the Gaussian renderer projects the propagated Gaussian distribution in a specific view for future scene reconstruction.

\subsection{Learning Objectives}\label{subsec:gaussian_world_model}

\noindent\textbf{Current Scene Consistency Loss.}
Reconstructing the current scene based on the current Gaussian parameters accurately can enhance the performance of the Gaussian regressor. To achieve this goal, we introduce the consistency objective between the realistic current observation and the rendered according to the current Gaussian parameters:
\begin{equation}\label{eq:objective_csc}
\mathcal{L}_{\text{\tiny Geo}}=\Vert \mathbf{C}^{(t)}-\hat{\mathbf{C}}^{(t)} \Vert_2^2,
\end{equation}where $\mathbf{C}^{(t)}$ and $\hat{\mathbf{C}}^{(t)}$ respectively mean the groundtruth and prediction of observation images from different views at the $t_{th}$ step.

\noindent\textbf{Semantic Feature Consistency Loss.}
The semantic features contain high-level visual information of the observed scenes. Since the foundation models can extract informative semantic features for general scenes, we expect the semantic features in our Gaussian parameters to mimic those acquired by large pre-trained models such as Stable Diffusion~\cite{rombach2022diffusion}, so that the knowledge learned by the pre-trained models can be distilled to our Gaussian world model according to the following objective:
\begin{equation}\label{eq:objective_sfc}
    \mathcal{L}_{\text{\tiny Sem}}=1-\sigma_{\cos}(\mathbf{F}^{(t)}, \hat{\mathbf{F}}^{(t)}),
\end{equation}where $\mathbf{F}^{(t)}$ and $\hat{\mathbf{F}}^{(t)}$ are projected map of semantic features in Gaussian parameters and the feature map learned by pre-trained models. $\sigma_{\cos}$ means the cosine distance between variables.

\noindent\textbf{Action Prediction Loss.}
The distribution parameters in our dynamic Gaussian framework are leveraged to predict the optimal action of the robot arm and grippers for general manipulation tasks. We employ a multi-modal transformer PerceiverIO~\cite{jaegle2021perceiver} to infer the selection probability of different action candidates based on the Gaussian parameters and the human language instructions, and leverage the cross-entropy loss $CE$ for accurate action prediction:
\begin{equation}\label{eq:objective_ap}
    \mathcal{L}_{\text{\tiny Act}}=CE(p_{\text{\tiny trans}}, p_{\text{\tiny rot}}, p_{\text{\tiny open}}, p_{\text{\tiny col}}),
\end{equation}
where $p_{\text{\tiny trans}}$, $p_{\text{\tiny rot}}$, $p_{\text{\tiny open}}$, $p_{\text{\tiny col}}$ represent the probability of the groundtruth actions in expert demonstrations for translation, rotation, gripper openness and collision avoidance of the robot, respectively.

\noindent\textbf{Future Scene Consistency Loss.}
We require consistency between the reconstructed and realistic scenes, so that the dynamic Gaussian Splatting framework can accurately embed scene-level spatiotemporal dynamics in the Gaussian parameters. Specifically, the training objective aligns the predicted future scenes based on different observations and actions with the realistic ones, which can be formulated as follows:
\begin{equation}\label{eq:world_model_loss}
\mathcal{L}_{\text{\tiny Dyna}}=\Vert \hat{\mathbf{C}}^{(t+1)}(a^{(t)}, o^{(t)}) - \mathbf{C}^{(t+1)} \Vert_2^2,
\end{equation}where $\hat{\mathbf{C}}^{(t+1)}(a^{(t)}, o^{(t)})$ means the predicted future image of the scene at the $t_{th}$ step based on the action $a^{(t)}$ and current observation $o^{(t)}$, and $\mathbf{C}^{(t+1)}$ is the realistic counterpart. 
By imposing the Gaussian world model to predict future scenes based on the representation, the representation is required to encode the physical properties of the scene. This is important for the action decoder to predict effective actions with such representation.

The overall objective for our \method agent is written as a weighted combination of different loss terms:
\begin{equation}
\mathcal{L}=\mathcal{L}_{\text{Act}} + \lambda_{\text{\tiny Geo}}\mathcal{L}_{\text{\tiny Geo}}+\lambda_{\text{\tiny Sem}} \mathcal{L}_{\text{\tiny Sem}}+\lambda_{\text{\tiny Dyna}} \mathcal{L}_{\text{\tiny Dyna}},
\end{equation}where $\lambda_{\text{\tiny Geo}}, \lambda_{\text{\tiny Sem}}, \lambda_{\text{\tiny Dyna}}$ are the hyperparamters that controls the importance of different terms during training. In training, we set a warm-up phase that freezes the deformation predictor to learn a stable representation model and a Gaussian regressor during the first \num{3}k iterations. After the warm-up phase, we jointly train the whole Gaussian world model with the action decoder.

%% file: sec/4_experiments.tex
\section{Experiments}
\label{sec:experiments}

In this section, we first introduce the experiment setup including datasets, baseline methods, and implementation details (Section \ref{subsec:ex_setup}). Then we compare our method with the state-of-the-art approaches to show the superiority in success rate (Section \ref{subsec:sota}), and conduct an ablation study to verify the effectiveness of different components in our dynamic Gaussian Splatting framework and the Gaussian world model (Section \ref{subsec:ablation}). Finally, we also illustrate the visualization results to depict our intuition (Section \ref{subsec:qualitative}). Further results and case studies can be found in the supplementary material.

\subsection{Experiment Setup}
\label{subsec:ex_setup}
\noindent\textbf{Simulation.} 
Our experiments are conducted in the popular RLBench~\cite{james2020rlbench} simulated tasks. Following~\cite{ze2023gnfactor}, we utilize a curated subset of $10$ challenging language-conditioned manipulation tasks from RLBench, which includes \num{166} variations in object properties and scene arrangement.
The diversity of these tasks requires the agent to acquire generalizable knowledge about the intrinsical scene-level spatial-temporal dynamics for manipulation, rather than solely relying on mimicking the provided expert demonstrations to achieve high success rates.
We evaluated 25 episodes in the testing set for each task to avoid result bias from noise.
For visual observation, we employ RGB-D images captured by a single front camera with a resolution of $128\times 128$. 
We use the same number of cameras (\ie, $20$) as GNFactor to provide multi-view supervision for fair comparisons.
During the training phase, we use \num{20} demonstrations for each task.

\noindent\textbf{Baselines:} We compare our \method with the previous state of the arts including the perceptive method PerAct~\cite{shridhar2023peract} and its modified version using \num{4} camera inputs to cover the workbench, as well as the generative method GNFactor~\cite{ze2023gnfactor}. The evaluation metric is the task success rate, which measures the percentage of completed episodes. An episode is considered successful if the agent completes the goal specified in natural language within a maximum of \num{25} steps.

\noindent\textbf{Implementation Details.} 
We use the $\text{SE}(3)$~\cite{shridhar2023peract,ze2023gnfactor} augmentation for the expert demonstrations in the training set to enhance the generalizability of agents.
To mitigate the impact of parameter size, we utilize the same version of PerceiverIO~\cite{jaegle2021perceiver} as the action decoder across all baselines.
All the compared methods are trained on two NVIDIA RTX \text{4090} GPUs for $100$k iterations with a batch size of $2$. We employ LAMB optimizer \cite{you2019lamb} with an initial learning rate \num{5e-4}. We also adopt a cosine scheduler with warmup in the first 3k steps.

\begin{table}[t]
\centering
\caption{\small \textbf{Multi-task Test Results.} We evaluate $25$ episodes per task for the final checkpoint on $10$ challenging tasks from RLBench and report the success rates (\%), where the second results are underlined and the best results are bold.
}
\label{table:comparison_with_sota} 
\setlength{\tabcolsep}{9.5pt}
\scriptsize
\renewcommand{\arraystretch}{1.1}%
\begin{tabular}{l*{5}{>{\centering\arraybackslash}p{32pt}}}
\toprule

 Method / Task  & \makecell{\texttt{close} \\ \texttt{jar}} & \makecell{\texttt{open} \\ \texttt{drawer}} & \makecell{\texttt{sweep to} \\ \texttt{dustpan}} & \makecell{\texttt{meat off} \\ \texttt{grill}} & \makecell{\texttt{turn} \\ \texttt{tap}} \\
 
\midrule

  PerAct & \dd{18.7} & \ddbfblue{54.7} & \dd{0.0} & \dd{40.0} & \dd{38.7} \\

  PerAct (4 cameras) & \dd{21.3} & \dd{44.0} & \dd{0.0} & \ddbf{65.3} & \dd{46.7} \\

 GNFactor & \ddbfblue{25.3} & \ddbf{76.0} & \ddbfblue{28.0} & \dd{57.3} & \ddbfblue{50.7}\\

 \method (ours) & \ddbf{28.0 } & \ddbf{76.0} & \ddbf{64.0} & \ddbfblue{60.0} & \ddbf{56.0}\\

\end{tabular}
\begin{tabular}{lccccc|c}
 \cmidrule{1-7}
  Method / Task & \makecell{\texttt{slide} \\ \texttt{block}} & \makecell{\texttt{put in} \\ \texttt{drawer}} & \makecell{\texttt{drag} \\ \texttt{stick}} & \makecell{\texttt{push} \\ \texttt{buttons}} & \makecell{\texttt{stack} \\ \texttt{blocks}} & \textbf{Average}\\
  
\cmidrule{1-7}

  PerAct & \dd{18.7} & \dd{2.7} & \dd{5.3} & \ddbfblue{18.7} & \ddbfblue{6.7} & $20.4$ \\

  PerAct (4 cameras) & \dd{16.0} & \ddbfblue{6.7} & \dd{12.0} & \dd{9.3} & \dd{5.3} & $22.7$\\

 GNFactor & \ddbfblue{20.0} & \dd{0.0} & \ddbfblue{37.3} & \ddbfblue{18.7} & \dd{4.0} & \ddbfblue{31.7} \\
 \method (ours) & \ddbf{24.0} & \ddbf{16.0} & \ddbf{92.0} & \ddbf{20.0} & \ddbf{12.0} & \ddbf{44.8} \\

\bottomrule
\end{tabular}
\end{table}

\begin{table}[t]
\centering
\caption{\small \textbf{Comparison of Methods with Different Techniques.}
Following~\cite{guhur2023instruction}, we manually group the \num{10} RLBench tasks into \num{6} categories according to their main challenges to demonstrate the improvement reason. The \num{6} categories are detailed in the supplementary file.}
\label{table:ablation_study}
\renewcommand{\arraystretch}{1.1}%
\setlength{\tabcolsep}{6pt}
\scriptsize
\begin{tabular}{ccc|cccccc|c}
\toprule
Geo. & Sem. & Dyna. & \texttt{Planning} & \texttt{Long} & \texttt{Tools} & \texttt{Motion} & \texttt{Screw} & \texttt{Occlusion} & \textbf{Average} \\
\midrule
\no & \no & \no & \dd{36.0} & \dd{2.0} & \dd{25.3} & \dd{52.0} & \dd{4.0} & \dd{28.0} & \dd{23.6} \\
\yes & \no & \no & \ddbfblue{46.0} & \dd{4.0} & \dd{52.0} & \dd{52.0} & \ddbfblue{24.0} & \dd{60.0} & \dd{39.2} \\
\yes & \yes & \no & \ddbfblue{46.0} & \dd{8.0} & \ddbfblue{53.3} & \ddbf{64.0} & \ddbf{28.0} & \dd{56.0} & \dd{41.6} \\

\yes & \no & \yes & \ddbf{54.0} & \ddbfblue{10.0} & \dd{49.3} & \ddbf{64.0} & \ddbfblue{24.0} & \ddbfblue{72.0} & \ddbfblue{43.6} \\

 \yes & \yes & \yes & \dd{40.0} & \ddbf{14.0} & \ddbf{60.0} & \ddbfblue{56.0} & \ddbf{28.0} & \ddbf{76.0} & \ddbf{44.8} \\
\bottomrule
\end{tabular}
\end{table}

\subsection{Comparison with the State-of-the-Art Methods}
\label{subsec:sota}
In this section, we compare our \method with previous state-of-the-art methods on the RLBench tasksuite. \Cref{table:comparison_with_sota} illustrates the comparison of the average success rate of each task.
Our method achieves the best performance with an average success rate of \num{44.8}\%, which is state-of-the-art, outperforming the previous arts including both perceptive and generative-based methods by a sizable margin.
The dominated generative-based method GNFactor leveraged a generalizable NeRF to learn informative latent representation for optimal action prediction, which showed effective improvement beyond the perceptive-based method PerAct.
However, it ignores the scene-level spatiotemporal dynamics that demonstrate the interaction among objects, and the predicted actions still fail to achieve human goals because of the incorrect interaction.
On the contrary, our \method learns the scene dynamics with the proposed dynamic Gaussian Splatting framework, so that the robotic agent can complete human instructions with accurate action prediction in unstructured environments.
As a result, our method outperforms the second-best GNFactor method by a relative improvement of \num{41.3}\%.
In the task \texttt{meat off grill} where the best performance was not reached, our method also ranks as second best. The experimental results illustrate the effectiveness of our proposed method across multiple language-conditioned robotic manipulation tasks.

\subsection{Ablation Study}
\label{subsec:ablation}

Our dynamic Gaussian Splatting framework models the propagation of diverse features in the Gaussian embedding space, and the Gaussian world model reconstructs the future scene according to the current scene by constraining the consistency between reconstructed and realistic scenes for dynamics mining.
We conduct an ablation study to verify the effectiveness of each presented component in \Cref{table:ablation_study}.
We first implement a vanilla baseline without any proposed technique, where we directly train the representation model and the action decoder to predict the robot actions. 
By adding the Gaussian regressor to predict the Gaussian parameters, the performance improves by \num{15.6}\% compared with the baseline.
\begin{wrapfigure}{r}{0.4\textwidth}
    \centering
    \includegraphics[width=0.4\textwidth]{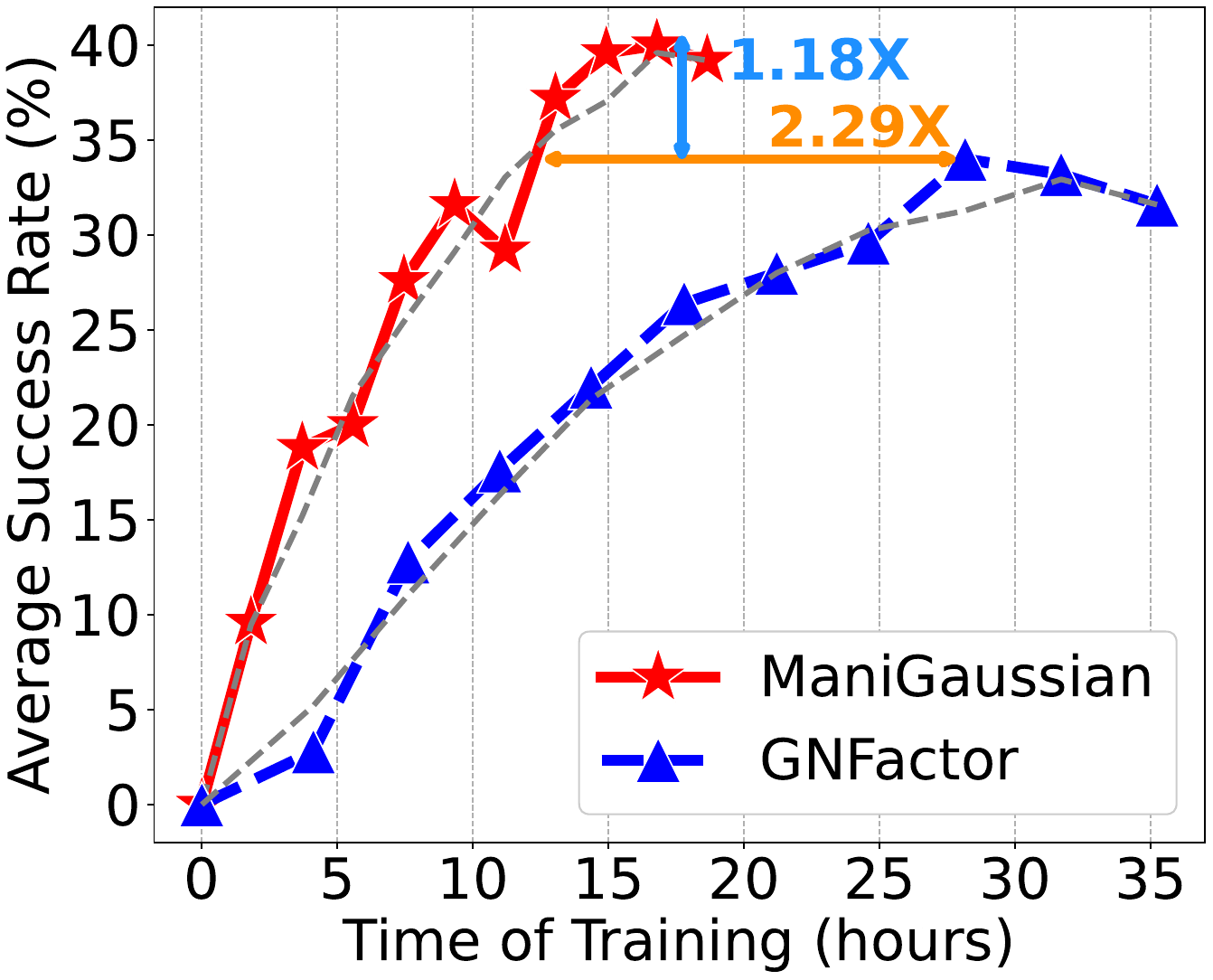}
    \caption{\small \textbf{Learning Curve.} Comparison of our \method with GNFactor in performance and speed. For a fair comparison, we exclude auxiliary losses from the reconstruction loss. The grey dotted lines represent the results using a moving average.}
    \label{fig:curve}
\end{wrapfigure}
Especially, in the tasks that require geometric reasoning such as \texttt{Occlusion}, \texttt{Tools} and \texttt{Screw}, it outperforms the vanilla version by sizable margins, which proves the ability of the Gaussian Splatting technology to model the spatial information for manipulation tasks.
We then add semantic features distilled from the pretrained foundation model into the dynamic Gaussian Splatting framework. By adding the semantic features and the related consistency loss, we observe that the average success rate increases by \num{2.4}\% than the only geometric features version, which indicates the benefits of the high-level semantic information for robotic manipulation.
Besides, we implement the deformation predictor and the corresponding future scene consistency loss, resulting in a dramatic performance improvement of \num{4.4}\%. Particularly, the proposed deformation predictor improves the task completion of \num{4} out of \num{6} task types, which demonstrates the importance of the scene-level dynamics encoded by the deformation predictor in the Gaussian world model, especially in long-horizon tasks (\texttt{Long}).
Though the dynamic loss may slightly impact short-term results due to the balance of different loss items, it significantly improves overall performance.
After combining all the techniques in our dynamic Gaussian Splatting framework, the performance increases from \num{23.6}\% to \num{44.8}\%, which verifies the necessity of the scene-level spatiotemporal dynamics mined by the proposed dynamics Gaussian Splatting framework with the Gaussian world model.

Figure \ref{fig:curve} depicts the learning curve of the proposed \method and the state-of-the-art method GNFactor, where we save checkpoints and test them every \num{10}k parameter updates.
Both the compared methods get convergence within \num{100}k training steps.
As shown in Figure \ref{fig:curve}, our \method outperforms the state-of-the-art method GNFactor, achieving \num{1.18}$\times$ better performance and \num{2.29}$\times$ faster training. This result proves that our \method not only performs better but also trains faster, which also shows the efficiency of the explicit Gaussian scene reconstruction than the implicit approach like NeRF.

\begin{figure}[t]
    \centering
    \includegraphics[width=1\textwidth]{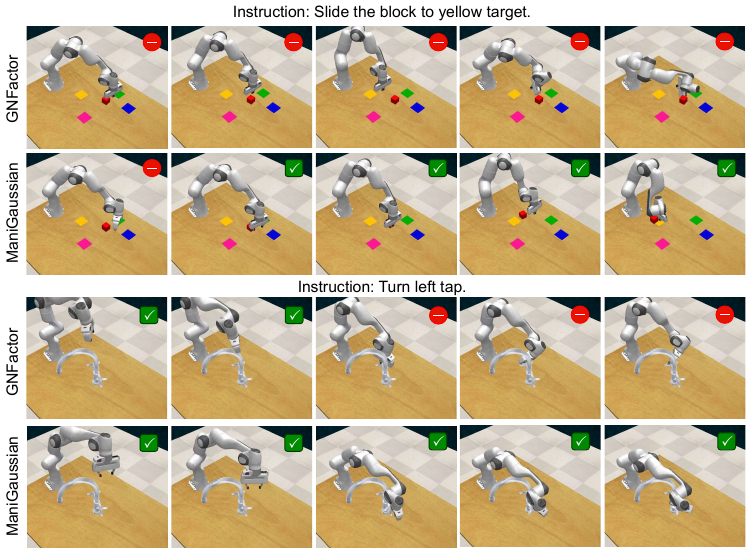}
    \caption{\small \textbf{Case Study.} The red mark signifies the pose deviates severely from the expert demonstration, whereas the green mark indicates that the pose aligns with the expert trajectory. Our \method can successfully complete the human goal with the physical understanding of scene-level spatial-temporal dynamics.}
    \label{fig:qualitative_case_study}
\end{figure}
\begin{figure}[t]
    \centering
    \includegraphics[width=1\textwidth]{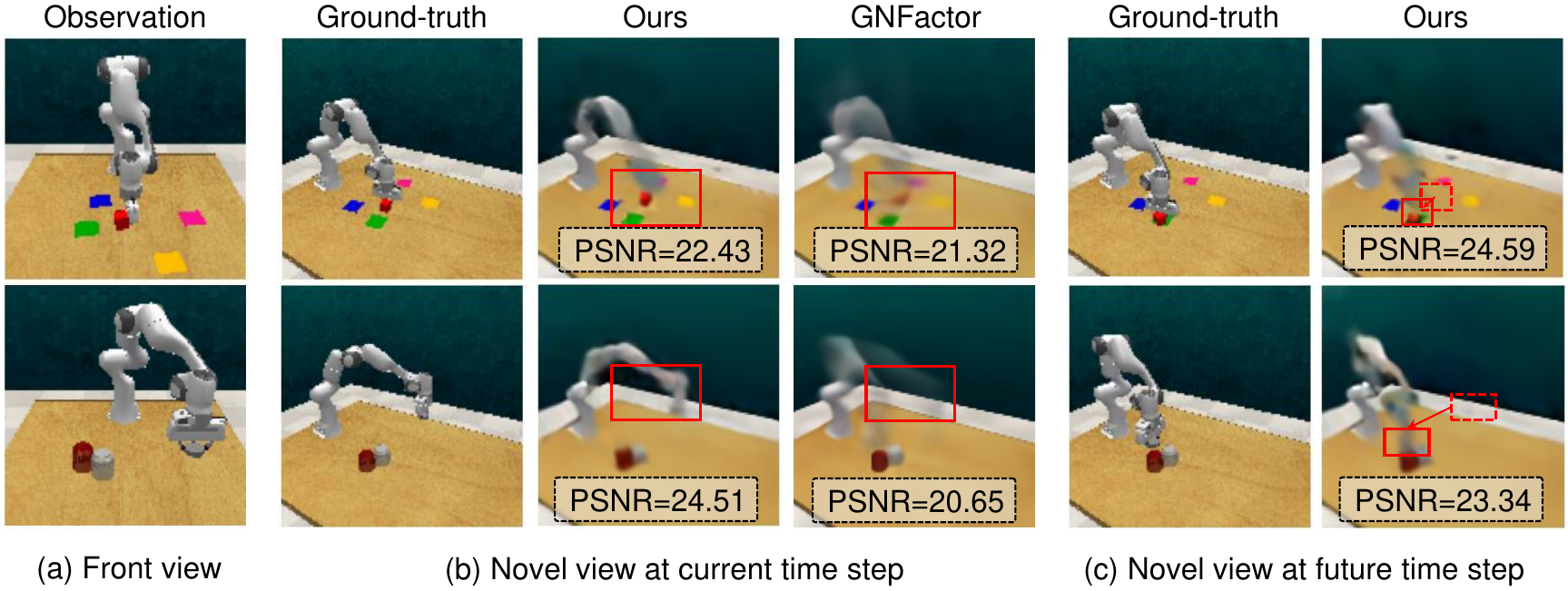}
    \caption{\small \textbf{Novel View Synthesis Results.} We remove the action loss here for better visualization. Our \method is capable of both current scene reconstruction and future scene prediction.}
    \label{fig:qualitative_novel_view_synthesis}
\end{figure}

\subsection{Qualitative Analysis}
\label{subsec:qualitative}
\noindent\textbf{Visualization of Whole Trajectories.}
We present two qualitative examples of the generated action sequence in \Cref{fig:qualitative_case_study} from GNFactor and our ManiGaussian. 
In the top case, the agent is instructed to \emph{``Slide the block to yellow target''}. The results show that the previous agent struggles to complete the task since it imitates the expert's backward pulling motion, even though the claw is already leaning towards the right side of the red block.
In contrast, \method returns to the red square and successfully slides the square to the yellow target, owing to that our method can correctly understand the scene dynamics of objects in contact. 
In the bottom case, the agent is instructed to \emph{``Turn left tap''}. The results show that GNFactor misunderstands the meaning of ``left'', and instead operates the right tap, and also fails to turn on the tap. In contrast, our ManiGaussian successfully completes the task, which shows that \method can not only understand the semantic information, but also execute operations accurately.

\noindent\textbf{Visualization of Novel View Synthesis.}
\Cref{fig:qualitative_novel_view_synthesis} shows the novel view image synthesis results.
First, based on the front view observation where the gripper shape cannot be seen, our ManiGaussian offers superior detail in modeling cubes in novel views. Second, our method accurately predicts future states based on the recovered details. For example, in the top case of the \texttt{slide block} task, our \method not only predicts the future gripper position that corresponds to the human instruction, but also predicts the future cube location influenced by the gripper based on the understanding of the physical interaction among objects. This qualitative result demonstrates that our \method learns the intricate scene-level dynamics successfully.

%% file: sec/5_conclusion.tex
\section{Conclusion}
\label{sec:conclusion}

In this paper, we have presented a \method agent that encodes the scene-level spatiotemporal dynamics for language-conditioned manipulation agents.
We design a dynamic Gaussian Splatting framework that models the propagation of features in the Gaussian embedding space, and the latent representation with scene dynamics is leveraged to predict the robot actions.
Subsequently, we build a Gaussian world model to parameterize the distributions in the dynamic Gaussian Splatting framework to mine scene-level dynamics by reconstructing the future scene.
Experiments in diverse manipulation tasks demonstrate the superiority of \method.
The limitations stem from the necessity of multiple view supervision with camera calibration for the Gaussian Splatting framework.

\section*{Acknowledgements}

This work was supported in part by the National Key Research and Development Program of China under Grant 2022ZD0114903 and Shenzhen Ubiquitous Data Enabling Key Lab under grant ZDSYS20220527171406015.
We would like to greatly thank Yanjie Ze for his kind response to the GNFactor issues.

%% file: sec/X_suppl.tex
% \clearpage
% \setcounter{page}{1}
\appendix
\maketitlesupplementary

In this supplementary material, we provide additional details and experiments not included in the main paper due to limitations in space.
\begin{itemize}
    \item \Cref{a:dataset}: Details of the RLBench dataset and the training pipeline used in our experiments.
    \item \Cref{a:implementation_details}: Additional implementation details of our \method.
    \item \Cref{a:additional_quantitative}: Supplementary quantitative analysis.
    \item \Cref{a:additional_qualitative}: Supplementary qualitative analysis.
\end{itemize}

\section{Details of RLBench}   % CPEM
\label{a:dataset}

\noindent\textbf{RLBench Dataset.}
\label{a:rlbench}
In this section, we provide a concise overview of the RLBench~\cite{james2020rlbench} dataset and our training pipeline.
Table~\ref{table:task desc} is an overview of the \num{10} selected tasks we use in the experiments. Our task variations include randomly sampled colors, sizes, counts, placements, and categories of objects. We have a color palette of \num{20} shades, including red, maroon, lime, green, blue, navy, yellow, cyan, magenta, silver, gray, orange, olive, purple, teal, azure, violet, rose, black, and white. 
The size of the objects is categorized into two types: short and tall. The number of objects can be either \num{1}, \num{2}, or \num{3}. 
Other properties vary depending on the specific task.
% For instance, there are \num{3} possible placement locations in \texttt{open drawer} task: top, middle and bottom. 
Furthermore, objects are randomly arranged on the tabletop within a certain range, adding to the diversity of the tasks.
In the ablation study, we adopt the task classification from~\cite{guhur2023instruction} to group the RLBench tasks of~\Cref{table:task desc} into \num{6} categories according to their key challenges. The task groups include:
\begin{itemize}
    \item The \texttt{Planning} group contains tasks with multiple subtasks. The included tasks are: \texttt{meat off grill} and \texttt{push buttons}.
    \item The \texttt{Long} group includes long-term tasks that requires more than \num{10} keyframes. The included tasks are: \texttt{put in drawer} and \texttt{stack blocks}.
    \item The \texttt{Tools} group requires the agent to grasp an object to interact with the target object. The included tasks are: \texttt{slide block}, \texttt{drag stick} and \texttt{sweep to dustpan}.
    \item The \texttt{Motion} group requires precise control, which often causes failures due to the predefined motion planner. The included task is: \texttt{turn tap}.
    \item The \texttt{Screw} group requires gripper rotation to screw an object. The included task is: \texttt{close jar}.
    \item The \texttt{Occlusion} group involves tasks with severe occlusion problems from certain views. The included task is: \texttt{open drawer}.
\end{itemize}
\begin{table}[htbp]
\vspace{-0.2in}
\caption{\textbf{Selected tasks.}}
\label{table:task desc}
\vspace{-0.1in}
\centering
\scriptsize
\setlength{\tabcolsep}{0.6 pt}
\begin{tabular}{lcccl}
\toprule
Task & Type & Variations & Keyframes & Instruction Template  \\ 
\midrule
\texttt{close jar} & color & 20 & 6.0 & \emph{``close the \underline{\hbox to 1mm{}} jar''} \\

\texttt{open drawer} & placement & 3 & 3.0 & \emph{``open the \underline{\hbox to 1mm{}} drawer''}\\

\texttt{sweep to dustpan} & size & 2 & 4.6 & \emph{``sweep dirt to the \underline{\hbox to 1mm{}} dustpan''} \\

\texttt{meat off grill} & category & 2 & 5.0 & \emph{``take the \underline{\hbox to 1mm{}} off the grill''} \\

\texttt{turn tap} & placement & 2 & 2.0 & \emph{``turn \underline{\hbox to 1mm{}} tap''}\\

\texttt{slide block} & color &  4 & 4.7 & \emph{``slide the block to \underline{\hbox to 1mm{}} target''} \\

\texttt{put in drawer} & placement & 3 & 12.0 & \emph{``put the item in the \underline{\hbox to 1mm{}} drawer''} \\

\texttt{drag stick} & color & 
20 & 6.0 &  \emph{``use the stick to drag the cube onto the \underline{\hbox to 1mm{}} target''}\\

\texttt{push buttons} & color & 50 &3.8 & \emph{``push the \underline{\hbox to 1mm{}} button, [then the \underline{\hbox to 1mm{}} button]''}
\\

\texttt{stack blocks} & color, count & 60 & 14.6 & \emph{``stack \underline{\hbox to 1mm{}} \underline{\hbox to 1mm{}} blocks''} \\

\bottomrule
\end{tabular}
\vspace{-0.2in}
\end{table}

\noindent\textbf{Training Pipeline.}
To learn the policy, we uniformly sample a group of expert episodes from all the task variations, and then randomly choose an input-action pair for each of the tasks to form a batch. Other sampling strategies (\eg, Auto-$\lambda$~\cite{liu2022auto}) are also available.
To simplify the tasks, the agent is assumed to access a predefined motion planner (\eg, RRT-Connect), so that the input-action pairs are determined as the bottleneck end-effector poses (\ie, keyframes) within each demonstration based on empirical rules: a pose is determined as a keyframe if the end-effector changes state (\eg close the gripper) or its velocities approach zero~\cite{chen2023polarnet,gervet2023act3d,james2022coarse,shridhar2023peract,ze2023gnfactor}.
% Under this setting, the action $\mathbf{a}^{(t)}$ at keyframe $t$ consists of translation $\mathbf{a}_{\text {trans}}^{(t)} \!\in\! \mathbb{R}^3$, rotation $\mathbf{a}_{\text {rot}}^{(t)} \in \mathbb{R}^{(360 / 5) \times 3}$, gripper open state $\mathbf{a}_{\text {open}}^{(t)} \in \mathbb{R}^2$, and the motion-planner state $\mathbf{a}_{\text {collide}}^{(t)} \in \mathbb{R}^2$.
This setting simplifies the sequential decision-making problem into predicting the next optimal keyframe action based on the current observation, which can also be interpreted as a classification task.

\section{Additional Implementation Details}
\label{a:implementation_details}

% ~\Cref{eq:gwm}
In this section, we detail the architectural design of each submodule in our Gaussian world model. For more details, please refer to our code.

\noindent\textbf{Representation model.}
The representation model $q_{\phi}$ is the same with~\cite{ze2023gnfactor}, which is not the main contribution in this paper. The representation model utilize a shallow 3D UNet to encode the voxel $\in\mathbb{R}^{100^3\times10}$ (RGB features, coordinates, indices, and occupancy) into the high-level visual features $\mathbf{v}^{(t)}\in \mathbb{R}^{100^3\times 128}$.

\noindent\textbf{Gaussian regressor.}
Given the current features $\mathbf{v}^{(t)}$ encoded by the representation model $q_{\phi}$, we pass it through a generalizable Gaussian regressor $g_{\phi}$ to infer the Gaussian distribution  $\theta^{(t)}$ directly.
The Gaussian regressor is designed as a lightweight multi-head neural network, where each head is responsible for predicting a specific feature. It consists of: (1) a position offset head that predicts the per-pixel 3D center offset  $\in\mathbb{R}^{3}$, (2) a color head that predicts the coefficients of the spherical harmonic basis $\in\mathbb{R}^{12}$, (3) a rotation head with normalization that predicts the rotation quaternion $\in\mathbb{R}^{4}$, (4) a scaling head with exponential activation that outputs the scaling factor $\in\mathbb{R}^{3}$, (5) an opacity head with sigmoid activation that predicts the opacity $\in\mathbb{R}^{1}$. (6) a semantic head that predicts the semantic feature $\in\mathbb{R}^{3}$.

\noindent\textbf{Deformation predictor.}
After obtaining the current visual features $\mathbf{v}^{(t)}$, Gaussian embedding $\theta^{(t)}$ and action $a^{(t)}$, we parameterize the transition process as a deformation predictor $p_\phi$ to predict the deformation $\Updelta \mu_{i}^{(t)} \in \mathbb{R}^{3}$ and $\Updelta r_{i}^{(t)} \in \mathbb{R}^{4}$ of each Gaussian, resulting in the future Gaussian embedding $\theta^{(t+1)}$. The deformation predictor is a fully-connected network with residual connections.

\noindent\textbf{Hyperparameters.}
The hyperparameters used in \method are shown in Table~\ref{table:hyperparam}. To train the robotic manipulation agent, we use $\lambda_{  \text{Geo}}=0.01$, $\lambda_{  \text{Sem}}=0.0001$ and $\lambda_{  \text{Dyna}}=0.001$ to focus on the action prediction. Other hyperparameters are in line with previous works~\cite{shridhar2023peract, ze2023gnfactor} for fair comparison.
\begin{table}[htbp]
\vspace{-0.2in}
\caption{\textbf{Hyperparameters}.}
\label{table:hyperparam}
\vspace{-0.1in}
\centering
\scriptsize
\begin{tabular}{cccccc}
\toprule
Hyperparameter & Value  \\ 
\midrule
training iteration & $100$k \\
image resolution & $128\times 128$\\
voxel resolution & $100\times 100 \times 100$\\
batch size & $2$ \\
optimizer & LAMB\\
learning rate & $0.0005$ \\
weight decay & $0.000001$ \\
Number of Gaussian points & $16384$ \\
$\lambda_{  \text{Geo}}$ & $0.01$ \\
$\lambda_{  \text{Sem}}$ & $0.0001$  \\
$\lambda_{  \text{Dyna}}$ & $0.001$  \\
\bottomrule
\end{tabular}
\vspace{-0.3in}
\end{table}

\section{Additional Quantitative Analysis}
\label{a:additional_quantitative}

We provide further ablation study on different implementation choices in our \method. \Cref{table:additional_ablation_study} presents the impact of different balance hyperparameters on the overall performance, from which we can conclude that the balance of each loss item is important to learn an optimal manipulation policy. 
\begin{table}[htbp]
\vspace{-0.2in}
\caption{\textbf{Impact of Balance Hyperparameters}.}
\label{table:additional_ablation_study}
\vspace{-0.1in}
\centering
\scriptsize
\setlength{\tabcolsep}{5pt}
\begin{tabular}{ccc|cccccc|c}
\toprule
$\lambda_{  \text{Geo}}$ & $\lambda_{  \text{Sem}}$ & $\lambda_{  \text{Dyna}}$ & \texttt{Planning} & \texttt{Long} & \texttt{Tools} & \texttt{Motion} & \texttt{Screw} & \texttt{Occlusion} & \textbf{Average} \\
\midrule
$0.01$ & $0$ & $0.00001$ & 42.0 & 24.0 & 48.0 &  48.0 & 28.0 & 72.0 & $42.4$ \\
$0.01$ & $0$ & $0.0001$ & 54.0 & 12.0 & 44.0 & 52.0 & 28.0 & 80.0 & $42.4$ \\
$0.01$ & $0$ & $0.001$ & $54.0$ & $10.0$ & $49.3$ & $64.0$ & $24.0$ & $72.0$ & $\textbf{43.6}$ \\
\midrule
$0.01$ & $0.00001$ & $0$ & 48.0 &  8.0  & 34.7 &   48.0  & 24.0   &     64.0 & $35.2$ \\
$0.01$ & $0.0001$ & $0$ & $46.0$ & $8.0$ & $53.3$ & $64.0$ & $28.0$ & $56.0$ & $\textbf{41.6}$ \\
$0.01$ & $0.001$ & $0$ & 46.0 & 2.0 & 37.3 &  60.0 & 40.0 &  68.0 & $37.6$ \\
\midrule
$0.01$ & $0.0001$ & $0.001$ & \dd{40.0} & $14.0$ & $60.0$ & $56.0$ & $28.0$ & $76.0$ & $\textbf{44.8}$ \\
\bottomrule
\end{tabular}
\vspace{-0.3in}
\end{table}

\section{Additional Qualitative Analysis}
\label{a:additional_qualitative}

We provide \num{9} additional comprehensive episodes generated by our \method and the state-of-the-art generative method GNFactor~\cite{ze2023gnfactor} in the attached video file (\emph{demo.mp4}). 
In the long-term \emph{``stack 2 rose blocks''}, \emph{``put the item in the bottom drawer''} and \emph{``take the steak off the grill''} tasks, \method completes the human instructions in the correct order with the understanding of high-level scene dynamics mined by the Gaussian world model. 
In the \emph{``sweep dirt to the short dustpan''} and \emph{``use the stick to drag the cube onto the azure target''} tasks that involve tool usage, our \method succeeds in solving the tasks by correctly understanding the low-level scene dynamics of objects in contact.
In the \emph{``slide the block to green target''}, \emph{``turn left tap''}, \emph{``close the azure jar''} and \emph{``open the bottom drawer''} tasks that require semantic understanding and precise control, our \method can successfully comprehend the semantic information to interact with the correct object instance, while the baseline method often confuses different instances.